\documentclass{llncs}
\usepackage[utf8]{inputenc}
\usepackage{numprint}
\usepackage{todonotes}
\usepackage{hyperref}
\usepackage{microtype}
\usepackage{listings}
\usepackage{wrapfig}
\usepackage{aliascnt}
\usepackage{times}
\usepackage{helvet}
\usepackage{courier}
\usepackage{verbatim}
\usepackage{hyperref}
\usepackage{booktabs}
\usepackage{siunitx}
\usepackage{graphicx}
\usepackage[vlined,linesnumbered,titlenumbered,ruled]{algorithm2e}
\usepackage[english]{babel}
\usepackage[utf8]{inputenc}
\usepackage{amsmath}
\usepackage{comment}
\usepackage{verbatim}
\usepackage{amsmath}
\usepackage{balance}
\usepackage{footnote}
\makesavenoteenv{tabular}
\usepackage{tablefootnote}

\usepackage{amssymb}
\usepackage[english]{babel}
\usepackage{epstopdf}
\usepackage{tabularx}
\usepackage{booktabs}
\usepackage{footmisc}
\usepackage{amsfonts}
\usepackage{graphicx}

\usepackage[T1]{fontenc}
\usepackage[scaled=0.85]{beramono}
\usepackage{listings}
\newcommand{\furl}[1]{\footnote{\scriptsize \url{#1}}}

\usepackage{enumitem}
\newlist{steps}{enumerate}{1}
\setlist[steps, 1]{label = Step \arabic*:}
\lstdefinelanguage{scala}{
  morekeywords={abstract,case,catch,class,def,%
    do,else,extends,false,final,finally,%
    for,if,implicit,import,match,mixin,%
    new,null,object,override,package,%
    private,protected,requires,return,sealed,%
    super,this,throw,trait,true,try,%
    type,val,var,while,with,yield},
  otherkeywords={=>,<-,<\%,<:,>:,\#,@},
  sensitive=true,
  morecomment=[l]{//},
  morecomment=[n]{/*}{*/},
  morestring=[b]",
  morestring=[b]',
  morestring=[b]"""
}
\usepackage{color}
\definecolor{dkgreen}{rgb}{0,0.6,0}
\definecolor{gray}{rgb}{0.5,0.5,0.5}
\definecolor{mauve}{rgb}{0.58,0,0.82}

\begin{document}
\title{EARL: Joint Entity and Relation Linking for \\Question Answering over Knowledge Graphs}
\author{Mohnish Dubey\inst{1,2}
 \and Debayan Banerjee\inst{1}
 \and Debanjan Chaudhuri\inst{1,2}
 \and  Jens Lehmann\inst{1,2}}
%
%
\authorrunning{Dubey, Banerjee, Chaudhuri, Lehmann}
\institute{
Smart Data Analytics Group (SDA), University of Bonn, Germany\\
\{dubey, chaudhur, jens.lehmann\}@cs.uni-bonn.de\\
\ debayan@uni-bonn.de
\and
Fraunhofer IAIS, Bonn, Germany\\
jens.lehmann@iais.fraunhofer.de
}

\maketitle
\begin{abstract}
Many question answering systems over knowledge graphs rely on entity and relation linking components in order to connect the natural language input to the underlying knowledge graph. 
Traditionally, entity linking and relation linking have been performed either as dependent sequential tasks or as independent parallel tasks. 
In this paper, we propose a framework called EARL, which performs entity linking and relation linking as a joint task. 
EARL implements two different solution strategies for which we provide a comparative analysis in this paper: 
The first strategy is a formalisation of the joint entity and relation linking tasks as an instance of the Generalised Travelling Salesman Problem (GTSP). In order to be computationally feasible, we employ approximate GTSP solvers.
The second strategy uses machine learning in order to exploit the connection density between nodes in the knowledge graph. It relies on three base features and re-ranking steps in order to predict entities and relations. 
We compare the strategies and evaluate them on a dataset with 5000 questions. 
Both strategies significantly outperform the current state-of-the-art approaches for entity and relation linking.

\end{abstract}


\section{Introduction}
Question answering over knowledge graphs (KGs) is an active research area concerned with techniques that allow obtaining information from knowledge graphs based on natural language input.
Specifically, Semantic Question Answering (SQA) as defined in~\cite{hoffner2017survey} is the task of users asking questions in natural language (NL) to which they receive a concise answer generated by a formal query over a KG. 
 
Semantic question answering systems can be a fully rule based systems~\cite{asknow2016dubey} or end-to-end machine learning based systems~\cite{serban2016generating}.
The main challenges faced in SQA are (i) entity identification and linking, (ii) relation identification and linking, (iii) query intent identification and (iv) formal query generation.

Some QA systems have achieved good performance on simple questions~\cite{lukovnikov2017neural}, i.e.~those questions which can be answered by linking to at most one relation and at most one entity in the KG.   
Recently, the focus has shifted towards complex questions~\cite{StagedYih2015semantic}, comprising of multiple entities and relations. 

Usually, all entities and relations need to be correctly linked to the knowledge graph in order to generate the correct formal query and successfully answer the question of a user. 
Hence, it is crucial to perform the linking process with high accuracy and this is a major bottleneck for the widespread adoption of current SQA systems.
In most entity linking systems~\cite{dbpedia2011mendes,agdistis2014usbeck}, disambiguation is performed by looking at other entities present in the input text. 
However, in the case of natural language questions (short text fragments) the number of other entities for disambiguation is not high. 
Therefore, it is potentially beneficial to consider entity and relation candidates for the input questions in combination, to maximise the usable evidence for the candidate selection process.
To achieve this, we propose EARL (\underline{E}ntity \underline{a}nd \underline{R}elation \underline{L}inker), a system for jointly linking entities and relations in a question to a knowledge graph. 
EARL treats entity linking and relation linking as a single task and thus aims to reduce the error caused by the dependent steps. 

EARL uses the knowledge graph to 
jointly disambiguate entity and relations:  
It obtains the context for entity disambiguation by observing the relations surrounding the entity. 
Similarly, it obtains the context for relation disambiguation by looking at the surrounding entities. 
The system supports multiple entities and relations occurring in complex questions. 
EARL implements two different solution strategies: 
The first strategy is a formalisation of the joint entity and relation linking tasks as an instance of the Generalised Travelling Salesman Problem (GTSP). Since the problem is NP-hard, we employ approximate GTSP solvers.
The second strategy uses machine learning in order to exploit the connection density between nodes in the KG. 
It relies on three base features and re-ranking steps in order to predict entities and relations. 
We compare the strategies and evaluate them on a dataset with 5000 questions. 
Both strategies outperform the current state-of-the-art approaches for entity and relation linking.

\begin{figure}[t]
	\centering
	\includegraphics[width=.95\linewidth]{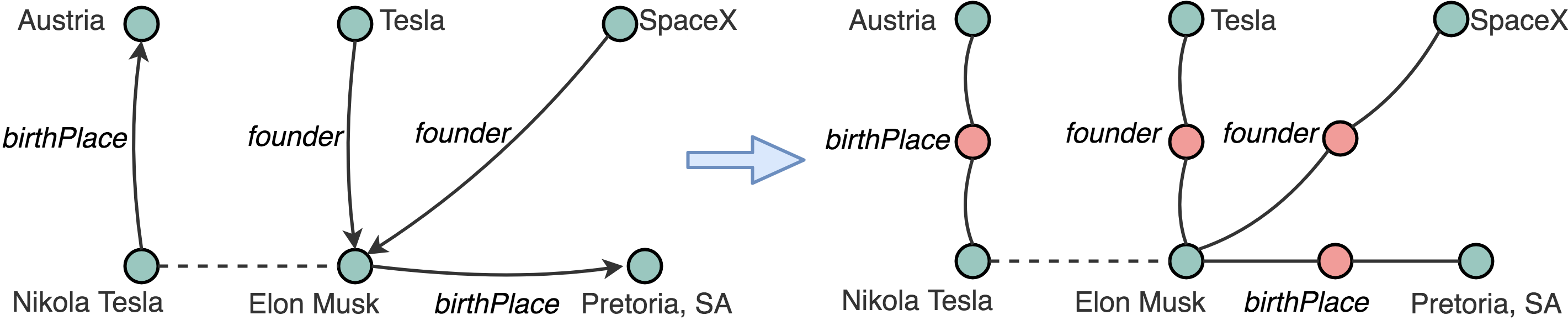}
	\caption{An excerpt of the subdivision knowledge graph for the example question "Where was the founder of Tesla and Space X born?". 
	Note that both entities and relations are nodes in the graph.}
	\label{fig:teslagraph} 
	\vspace{-15pt}
\end{figure}

Let us consider an example to explain the underlying idea: \textit{"Where was the founder of Tesla and SpaceX born?"}. 
Here, the entity linker needs to perform disambiguation for the keyword "Tesla" between the scientist "Nikola Tesla" and the car company "Tesla Motors". 
EARL uses all other entities and relations (\textit{SpaceX, founder, born}) present in the query.  
It does this by analysing the subdivision graph of the knowledge graph fragment containing the candidates for relevant entities and relations. 
While performing the joint analysis (Figure~\ref{fig:teslagraph}), EARL detects that there is no likely combination of candidates, which supports the disambiguation of "Tesla" as "Nikola Tesla", whereas there is a plausible combination of candidates for the car company "Tesla Motors". 

Overall, our contributions in this paper are as follows:
\begin{enumerate}	
\item The framework EARL, where GTSP solver or Connection Density can be used for joint linking of entities and relations (Sec.~\ref{earl}).
\item A formalisation of the joint entity and relation linking problem as an instance of the Generalised Travelling Salesman (GTSP) problem (Sec.~\ref{gtsp}).
\item An implementation of the GTSP strategy using approximate GTSP solvers. 
\item A "Connection Density" formalisation and implementation of the joint entity and relation linking problem as a machine learning task (Sec.~\ref{cd}).
\item An adaptive E/R learning module, which can correct errors occurring across different modules (Sec.~\ref{ad-er}). 
\item A comparative analysis of both strategies - GTSP and connection density (Table~\ref{cd-vs-gtsp}).
\item A fully annotated version of the 5000 question LC-QuAD data-set, where entity and relations are linked to the KG.
\item A large set of labels for DBpedia predicates and entities covering the syntactic and semantic variations.\footnote{dataset available at \url{https://github.com/AskNowQA/EARL}}
\end{enumerate}

The paper is organised into the following sections: 
(2) Related Work outlining some of the major contributions in entity and relation linking used in question answering;
(3) Problem Statement, where we discuss the problem in depth and our hypotheses for the solution;  
(4) the architecture of EARL 
including preprocessing steps followed by (i) a GTSP solver or (ii) a connection density approach; 
(5) Evaluation, with various evaluation criteria and results; (6) Discussion; and 
(7) Conclusion.

\section{Related Work} \label{relatedWork}

The entity and relation linking challenge has attracted a wide variety of solutions over time.  
Linking natural language phrases to DBpedia resources, Spotlight~\cite{dbpedia2011mendes} breaks down the process of entity spotting into four phases. 
It identifies the entity using a list of surface forms and then generates DBpedia resources candidates. 
It then disambiguates the entity based on surrounding context. 
AGDISTIS~\cite{agdistis2014usbeck} follows the inherent structure of the target knowledge base more closely to solve the problem. 
Being a graph-based disambiguation system, AGDISTIS performs disambiguation based on the hop-distance between the candidates for the entities in a given text, where multiple entities are present. 
Babelfy~\cite{babelfy} uses word sense disambiguation for entity linking. 
On the other hand, S-MART~\cite{smart2015Yang} is often appropriated as an entity linking system over Freebase resources. 
It generates multiple regression trees and then applies sophisticated structured prediction techniques to link entities to resources. 

As relation linking is generally considered to be a problem-specific task, only a few general purpose relation linking systems are in use.  
Iterative bootstrapping strategies for extracting RDF resources from unstructured text have been explored in BOA~\cite{boagerber2011bootstrapping} and PATTY~\cite{nakashole2012patty}. 
It consists of natural language patterns corresponding to relations present in the knowledge graph.
Word embedding models are also frequently used to overcome the linguistic gap for relation linking. 
RelMatch~\cite{ReMatchKuldeep} improves the accuracy of the PATTY dataset for relation linking. 
There are tools such as ReMatch~\cite{RelMatch} which uses wordnet similarity for relation linking.

Many QA systems use an out-of-the-box entity linker, often one of the aforementioned ones. 
These tools are not tailor-made for questions and are instead trained on large text corpora, typically devoid of questions. 
This may create several problems as questions do not span over more than one sentence, thereby rendering context-based disambiguation relatively ineffective. 
Further, graph based systems rely on the presence of multiple entities in the source text and disambiguate them based on each other. This becomes difficult when dealing with questions, as they 
seldom consist of multiple entity.

\begin{table*}[t]
\label{my-label-rel}
\begin{tabular}{p{1.5cm} p{1.8cm} p{4cm} p{4cm}}  \hline

\toprule
\bfseries{Linking Approach}    & \bfseries{QA System} & \bfseries{Advantage} & \bfseries{Disadvantage} \\ 
\toprule
\textbf{Sequential} & \cite{asknow2016dubey} \cite{both2016qanary}\cite{singh2018reinvent}   & -Reduces candidate search space for Relation Linking      & -Relation Linking information cannot be exploited in Entity Linking process\\
 & & -Allows schema verification  & - Errors in Entity Linking cannot be overcome
\\  \hline

\textbf{Parallel}   & \cite{veyseh2016cross} 
\cite{xserxu2014answering} 
\cite{park2015isoft} & - Lower runtime &  - Entity Linking process cannot use information from Relation Linking process and vice versa \\
 & & - Re-ranking of Entities possible based on Relation Linking & - Does not allow schema verification
\\  \hline

\textbf{Joint}  & \cite{berant2013semantic} \cite{StagedYih2015semantic} 
  & -  Potentially high accuracy  & - Complexity increase  \\      
(with & & - Reduces error propagation & - Larger search space  \\
limited   & & - Better disambiguation & \\
candidate  & & - Allows schema verification  &   \\      
set) & & - Allows re-ranking\\ 
\bottomrule

\end{tabular}

\caption{State of the art for Entity and Relation linking in Question Answering} 
      \vspace{-15pt}
\label{tab:er_linking}
\vspace{-14.5pt}
\end{table*}

Thus, to avoid the issues mentioned, a variety of approaches have been employed for entity and relation linking for question answering.
Semantic parsing-based systems such as AskNow~\cite{asknow2016dubey} and TBSL~\cite{tbslunger2012template} first link the entities and generate a list of candidate relations based on the identified resources. 
They use several string and semantic similarity techniques to finally select the correct entity and relation candidates for the question. 
In these systems, the process of relation linking depends on linking the entities. 
Generating entity and relation candidates has also been explored by~\cite{StagedYih2015semantic}, which uses these candidates to create staged query graphs, and later re-ranks them based on textual similarity between the query and the target question, computed by a Siamese architecture-based neural network.
There are some QA systems such as Xser~\cite{xserxu2014answering}, which performs relation linking independent of entity linking.
STAGG\cite{StagedYih2015semantic} takes the top 10 entities given by the entity linker and tries to build query-subgraph chains corresponding to the question. This approach considers a ranked list of entity candidates from the entity linker and chooses the best candidate based on the query subgraph formed.
Generally, semantic parsing based systems treat entity and relation linking as separate tasks which can be observed in the generalised pipeline of Frankenstein~\cite{singh2018reinvent} and OKBQA~\url{www.okbqa.org/}.

\section{Overview and Preliminaries}

\subsection{Overview and Research Questions}

As discussed previously, in question answering the tasks of entity and relation linking are performed either sequentially or in parallel. 
In sequential systems, usually the entity linking task is performed first, followed by relation linking. 
As a consequence, information in the relation linking phase cannot be exploited during entity linking in this case. 
In parallel systems, entity and relation linking are performed independently. 
While this is efficient in terms of runtime performance, the entity linking process cannot benefit from further information obtained during relation linking and vice versa.
We illustrate the advantages and disadvantages of both approaches, as well as the systems following them, in Table~\ref{tab:er_linking}.
Our main contribution in this paper is the provision of a system, which takes candidates for entity and relation linking as input and performs a joint optimisation selecting the best combination 
of entity and relation candidates.
\paragraph{Postulates}

We have three postulates, which we want to verify based on our approach: 

\textbf{H1}: Given candidate lists of entities and relations from a question, the correct solution is a cycle of minimal cost that visits exactly one candidate from each list.

\textbf{H2}: Given candidate lists of entities and relations from a question, the correct candidates exhibit relatively dense and short-hop connections among themselves in the knowledge graph compared to wrong candidate sets.

\textbf{H3}: Jointly linking entity and relation leads to higher accuracy compared to performing these tasks separately.

We will re-visit all of these postulates in the evaluation section of the paper.

\subsection{Preliminaries}

We will first introduce basic notions from graph theory:

\begin{definition}[Graph]
A (simple, undirected) graph is an ordered pair $G = (V,E)$ where $V$ is a set whose elements are called \emph{vertices} and $E$ is a set of 
pairs of vertices which is called \emph{edges}.
\end{definition}

\begin{definition}[Knowledge Graph]
Within the scope of this paper, we define a \emph{knowledge graph} as a labelled directed multi-graph. A labelled directed multi-graph is a tuple $KG = (V, E, L)$ where $V$ is a set called vertices,  $L$ is a set of edge labels and $E \subseteq V \times L \times V$ is a set of ordered triples. 
\label{def:kg}
\end{definition}

It should be noted that our definition of knowledge graphs captures basic aspects of RDF datasets as well as property graphs~\cite{gubichev2014graph}. The knowledge graph vertices represent entities and the edges represent relationships between those entities.

\begin{definition}[Subdivision Graph]
The subdivision graph  \cite{trudeau1993introduction} $S(G)$ of a graph $G$ is the graph obtained from $G$ by replacing each edge $e = (u,v)$ of $G$ by a new vertex $w_e$ and 2 new edges $(u,w_e)$ and $(v,w_e)$ .
\label{def:subdiv}
\end{definition}

\label{sec:architecture}

\section{EARL} \label{earl}

In general, entity linking is a two step process. 
The first step is to identify and spot the span of the entity. 
The second step is to disambiguate or link the entity to the knowledge graph. 
For linking, the candidates are generated for the spotted span of the entity and then the best candidate is chosen for the linking.
These two steps are similarly followed in standard relation linking approaches.
In our approach, we first spot the spans of entities and relations. 
After that, the (disambiguation) linking task is performed jointly for both entities and relations.

In this section we first discuss the step of span detection of entity and relation in natural language question and candidate list generation.
We perform the disambiguation by two different approaches, which are discussed later in this section.

\begin{figure*}[t!]
	\centering
	\includegraphics[width=0.77\linewidth]{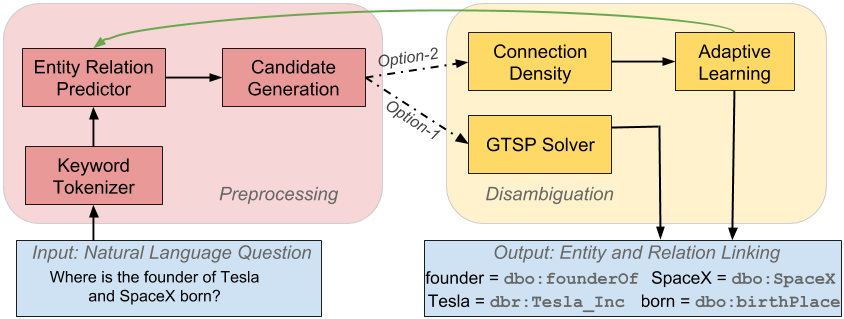}
	\caption{EARL Architecture: In the disambiguation phase one may choose either Connection Density or GTSP. In cases where training data is not available beforehand GTSP works better.}
	\label{fig:WorkFlow} 
	\vspace{-15pt}
\end{figure*}

\subsection{Candidate Generation Steps}

\subsubsection{Shallow Parsing}: Given a question, extract all keyword phrases out. 
EARL uses SENNA\cite{collobert2011natural} as the keyword extractor.
We also remove stop words from the question at this stage. 
In example question "Where was the founder of Tesla and SpaceX born?" we identify \textit{<founder, Tesla, SpaceX, born>} as our keyword phrases.

\subsubsection{E/R Prediction} Once keyword phrases are extracted from the questions, the next step in EARL is to predict whether each of these is an entity or a relation. 
We use a character embedding based long-short term memory network (LSTM) to do the same. The network is trained using labels for entity and relation in the knowledge graph. 
For handling out of vocabulary words~\cite{pinter2017mimicking}, and also to encode the knowledge graph structure in the network, we take a multi-task learning approach with hard parameter sharing. 
Our model is trained on a custom loss given by:

\begin{equation}{ \mathcal{E} = (1 - \alpha) * \mathcal{E}_{BCE} + \alpha * \mathcal{E}_{ED} } \end{equation}

Where, \begin{math} \mathcal{E}_{BCE} \end{math} is the binary cross entropy loss for the learning objective of a phrase being an entity or a relation and \begin{math} \mathcal{E}_{Ed} \end{math} is the squared eucledian distance between the predicted embedding and the correct embedding for that label. The value of \begin{math}\alpha \end{math} is empirically selected as 0.25.
We use pre-trained label embeddings from RDF2Vec~\cite{ristoski2016rdf2vec} which are trained on knowledge graphs. RDF2Vec provides latent representation for entities and relations in RDF graphs. It efficiently captures the semantic relatedness between entities and relations.

We use a hidden layer size of 128 for the LSTM, followed by two dense layers of sizes 512 and 256 respectively. A dropout value of 0.5 is used in the dense layers. The network is trained using Adam optimizer~\cite{kingma2014adam} with a learning rate of 0.0001 and a batch size of 128.
Going back to the example, this module identifies "founder" and "born" as \textit{relations}, "Tesla" and "SpaceX" as \textit{entities}.

\subsubsection{Candidate List Generation}
This module retrieves a candidate list for each keyword identified in the natural language question by the shallow parser. To retrieve the top candidates for a keyword we create an Elasticsearch\footnote{~\url{https://www.elastic.co/products/elasticsearch}} index of URI-label pairs. 
Since EARL requires an exhaustive list of labels for a URI in the knowledge graph, we expand the labels. 
We used Wikidata labels for entities which are in same-as relation in the knowledge base.
For relations we require labels which were semantically equivalent (such as writer, author) for which we took synonyms from the Oxford Dictionary API \footnote{~\url{https://developer.oxforddictionaries.com/}}.
To cover grammatical variations of a particular label, we added inflections from fastText\footnote{~\url{https://fasttext.cc/}}.
We avoid any bias held towards or against popular entities and relations.

The output of these pre-processing steps are i) set of keywords from the question, ii) every keyword is identified either as relation or entity, iii) for every keyword there is a set of candidate URIs from the knowledge graph.

\subsection{Using GTSP for Disambiguation} \label{gtsp}

At this point we may use either a GTSP based solution or Connection Density (later explained in 4.3) for disambiguation. We start with the formalisation for GTSP based solution.

\begin{figure}[b]
\vspace{-5pt}
	\centering
	\includegraphics[width=.9\linewidth]{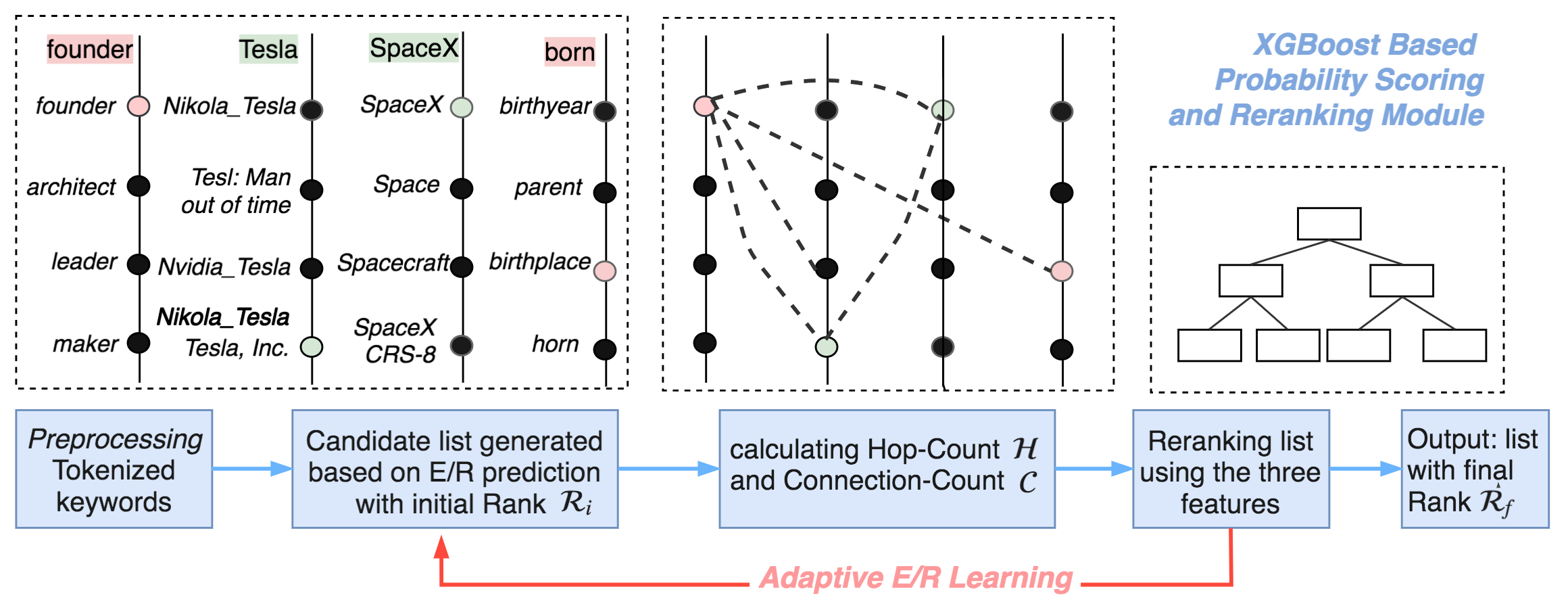}
\vspace{-10pt}
	\caption{Connection Density with example: The dotted lines represent corresponding connections between the nodes in the knowledge base.}
	\label{fig:connectiondensity} 
	\vspace{-10pt}
\end{figure}

The entity and relation linking process can be formalised via spotting and candidate generation functions as follows: 
Let $S$ be the set of all strings. We assume that there is a function $spot : S \to 2^S$ which maps a string $s$ (the input question) to a set $\mathcal{K}$ of substrings of $s$. We call this set $\mathcal{K}$ the \emph{keywords} occurring in our input.
Moreover, we assume there is a function $\text{cand}_{KG}: \mathcal{K} \to 2^{V \cup L}$ which maps each keyword to a set of candidate node and edge labels for our knowledge graph $G = (V, E, L)$.
The goal of joint entity and relation linking is to find combinations of candidates, which are closely related. How closely nodes are related is modelled by a cost function $\text{cost}_{KG}: (V \cup L) \times (V \cup L) \to [0,1]$. Lower values indicate closer relationships.
According to our first postulate, we aim to encode graph distances in the cost function to reward those combinations of entities and relations, which are located close to each other in the input knowledge graph.  
To be able to consider distances between both relations and entities, we transform the knowledge graph into its subdivision graph (see Definition~\ref{def:subdiv}). 
This subdivision graph allows us to elegantly define the distance function as illustrated in Figure~\ref{fig:connectiondensity}.


Given the knowledge graph $KG$ and the functions $spot$, $cand$ and $cost$, we can cast the problem of joint entity and relation linking as an instance of the Generalised Travelling Salesman (GTSP) problem:
We construct a graph $G$ with $V = \bigcup_{k \in K} \text{cand}(k)$. Each node set $\text{cand}(k)$ is called a cluster in this vertex set. The GTSP problem is to find a subset $V'  = (v_1, \dots, v_n)$ of $V$ which contains exactly one node from each cluster and the total cost $\sum_{i=1}^{n-1} \text{cost}(v_i,v_{i+1})$ is minimal with respect to all such subsets. Please note that in our formalisation of the GTSP, we do not require $V'$ to be a cycle, i.e.~$v_1$ and $v_n$ can be different. 
Moreover, we do not require clusters to be disjoint, i.e.~different keywords can have overlapping candidate sets.

\begin{figure}[t]
	\centering
	\includegraphics[width=.75\linewidth]{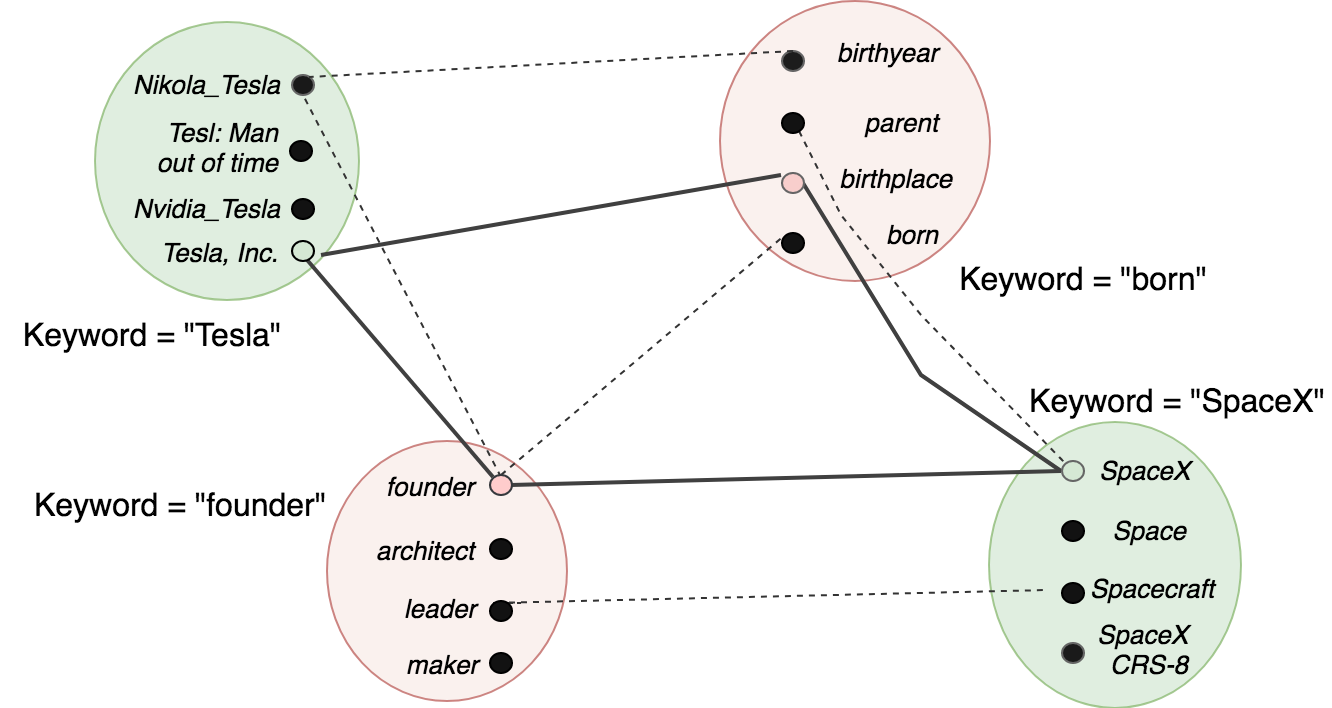}
	\caption{Using GTSP for disambiguation : The bold line represents the solution offered by the GTSP solver. Each edge represents an existing connection in the knowledge graph. The edge weight is equal to the number of hops between the two nodes in the knowledge graph. We also add the index search ranks of the two nodes the edges connect to the edge weight when solving for GTSP. }
	\label{fig:gtsp} 
	\vspace{-15pt}
\end{figure}

Figure~\ref{fig:gtsp} illustrates the problem formulation. 
Each candidate set for a keyword forms a cluster in the graph. 
The weight of each edge in this graph is given by the cost function, which includes the distance between the nodes in the subdivision graph of the input knowledge graph as well as the confidence scores of the candidates.  
The GTSP requires the solution to visit one element per cluster and minimises the overall distance.

\subsubsection*{Approximate GTSP Solvers}
In order to solve the joint entity and relation linking problem, the corresponding GTSP instance needs to be solved. 
Unfortunately, the GTSP  is NP-hard~\cite{LAPORTE1987185} and hence it is intractable. However, since GTSP can be reduced to standard TSP, several polynomial approximation algorithms exist to solve GTSP. 
The state-of-the-art approximate GTSP solver is the Lin–Kernighan–Helsgaun algorithm~\cite{helsgaun2015solving}.
Here, a GTSP instance is transformed into standard asymmetric TSP instances using the Noon-Bean transformation.
It allows the heuristic TSP solver LKH to be used for solving the initial GTSP.
Among LKH's characteristics, its use of 1-tree approximation for determining a candidate edge set, the extension of the basic search step, and effective rules for directing and pruning the search contribute to its efficiency.

While a GTSP based solution would be suitable for solving the joint entity and relation linking problem, it has the drawback that it can only provide the best candidate for each keyword given the list of candidates. Most approximate GTSP solutions do not explore all possible paths and nodes and hence a comprehensive scoring and re-ranking of nodes is not possible.
Ideally, we would like to go beyond this and re-rank all candidates for a given keyword. 
This would open up new opportunities from a QA perspective, i.e.~a user could be presented with a sorted list of multiple possible answers to select from.

\subsection{ Using Connection Density for Disambiguation} \label{cd}

\begin{table*}
\vspace{-15pt}
\label{my-label}
\begin{tabular}{p{5.9cm} p{0.1cm} p{5.9cm}}  \hline

\toprule
\bfseries{GTSP}& & \bfseries{Connection Density}  \\  
\toprule
 Requires no training data 
 &  & Requires data to train the XGBoost classifier
\\  \hline
The approximate GSTP LKH solution is only able to return the top result as not all possible paths are explored. & & Returns a list of all possible candidates in order of score
\\  \hline
Time complexity of LKH is $\mathcal{O}(n L^2)$ where n = number of nodes in graph, L = number of clusters in graph  of    &  & Time complexity is $\mathcal{O}(N^2 L^2)$ where N = number of nodes per cluster, L = number of clusters in graph
\\  \hline
Relies on identifying the path with minimum cost   &  & Depends on identifying dense and short-hop connections
\\  \bottomrule

\end{tabular}

\caption{Comparison of GTSP based approach and Connection density for Disambiguation }

 \vspace{-10pt}
\label{cd-vs-gtsp}
\end{table*}
\vspace{-15pt}

As discussed earlier, once the candidate list generation is achieved, EARL offers two independent modules for the entity and relation linking.  
In the previous subsection 4.2 we discussed one approach using GTSP.
In this subsection we will discuss the second approach for disambiguation using Connection Density, which works as an alternative to the GTSP approach.
We have also compared the two methods in table 2.
\subsubsection{Formalisation of Connection Density:}
For identified keywords in a question we have the set $\mathcal{K}$ as defined earlier. 
For each keyword $K_i$ we have list $L_i$ which consists of all the candidate uris generated by text search.
We have $n$ such candidate lists for each question given by, $\mathcal{L} = \{ L_1, L_2, L_3,... , L_n \}$.
We consider a probable candidate $c_m^i$ $\in L_i$, where $m$ is the total number of candidates to be considered per keyword, which is the same as the number of items in each list.

The hop distance $dKGhops(c_i^k ,c_j^o)$ $\in \mathbb{Z^+}$ is number of hops between $c_i^k$ and $c_j^o$ in the subdivision knowledge graph. 
If the shortest path from $c_i^k$ and $c_j^o$ requires the traversal of $h$ edges then $dKGhops(c_i^k ,c_j^o)$ = $h$. 

Connection Density is based on the three features: Text similarity based initial Rank of the List item ($\mathcal{R}_i$)  Connection-Count ($\mathcal{C}$)  and Hop-Count ($\mathcal{H}$)

Initial Rank of the List ($\mathcal{R}_i$), is generated by retrieving the candidates from the search index via text search. 
This is achieved in the preprocessing steps as mentioned in the section 4.
Further, to define $\mathcal{C}$ we introduce $dConnect$.


\begin{equation}
       dConnect(c_i^k ,c_j^o) = 
        \begin{cases}
            1 & \text{if $dKGhops(c_i^k$, $c_j^o)  	\leqslant 2$ } \\
            0 & \text{otherwise}
        \end{cases}
    \end{equation}

The Connection-Count $\mathcal{C}$ for an candidate $c$, is the number of connections from $c$ to candidates in all the other lists divided by the total number $n$ of keywords spotted. We consider nodes at hop counts of greater than 2 disconnected because nodes too far away from each other in the knowledge base do not carry meaningful semantic connection to each other.
\begin{equation}
        \mathcal{C}(c_i^k) = 1/n \sum_{o|o \neq k} \sum_{j=1}^{j = m} dConnect(c_i^k, c_j^o) 
\end{equation}

The Hop-Count $\mathcal{H}$ for a candidate $c$, is the sum of distances from $c$ to all the other candidates in all the other lists divided by the total number of keywords spotted.

\begin{equation}
    \mathcal{H}(c_i^k) = 1/n \sum_{o|o \neq k}\sum_{j=1}^{j = m} dKGhops(c_i^k, c_j^o)  
\end{equation}

\subsubsection{Candidate Re-ranking:}
$\mathcal{H}, \mathcal{C}$ and $\mathcal{R}_i$ constitute our feature space $\mathcal{X}$.
This feature space is used to find the most relevant candidate given a set of candidates for an identified keyword in the question. We use a machine learning classifier to learn the probability of being the most suitable candidate $\Bar{\textbf{c}}^i$ given the set of candidates. 
The final list $\mathcal{R}_f$ is obtained by re-ranking the candidate lists based on the probability assigned by the classifier. Ideally, $\Bar{\textbf{c}}^i$ should be the top-most candidate in $\mathcal{R}_f$.

The training data consists of the features $\mathcal{H}, \mathcal{C}$ and $\mathcal{R}_i$ and a label 1 if the candidate is the correct, 0 otherwise. For the testing, we apply the learned function from the classifier ${f}$ on $\mathcal{X}$ for every candidate $\in c_i$ and get a probability score for being the most suitable candidate.
We perform experiments with three different classifiers, namely extreme gradient boosting(xgboost), SVM(with a linear kernel) and logistic regression to re-rank the candidates. The experiments are done using a 5-fold cross-validation strategy where, for each fold we train the classifier on the training set and observe the mean reciprocal rank (MRR) of $\Bar{\textbf{c}}^i$ on the testing set after re-ranking the candidate lists based on the assigned probability. 
The average MRR on 5-fold cross-validation for the three classifiers are 0.905, 0.704 and 0.794 respectively. Hence, we use xgboost as the final classifier in our subsequent experiments for re-ranking.

\subsubsection{Algorithm }
We now present a pseudo-code version of the algorithm to calculate the two features:
Connection Density algorithm is used for finding hop count and connection count for each candidate node. We then pass these features to a classifier for scoring and ranking
This algorithm (Algorithm 1 Connection Density) has a time complexity given by $\mathcal{O}(N^2L^2)$ where N is the number of keywords and L is the number of candidates for each keyword.

\vspace{-15pt}
\begin{algorithm}

\SetKwInOut{Input}{input} \SetKwInOut{Output}{output}
\SetKwInOut{Fun}{function}
  \Fun{ConnectionDensity( )}
  \Input{$\mathcal{L}$ , with $n$ number of keywords \tcp{an array of arrays}}
  \Output{Hop-Count $\mathcal{H}$, Connection-Count $\mathcal{C}$}
 
  dConnectCounter = \{ \} \tcp{Count for connections from and to each node}
  dHopCounter = \{ \} \tcp{Similarly hop counts for each node}
  \ForEach{$L_a \in \mathcal{L}$}{
   \ForEach{$c_i^a \in L_a$}{ 
   	$dConnectCounter$[$c_i^a$] = 0  \tcp{Initialising the dictionary}
   	$dHopCounter$[$c_i^a$] = 0
   }
  }
  \ForEach{$(L_a,L_b) \in \mathcal{L}$}{
   \ForEach{$c_i^a \in L_a$}{
   	\ForEach{$c_j^b \in L_b$}{
    \If{$dKGhops(c_i^a$,$c_j^b$) <= 2} {
    $dConnectCounter$[$c_i^a$] += 1 \\
    $dConnectCounter$[$c_j^b$] += 1 \\
    }
    $dHopCounter$[$c_i^a$] += dKGhops($c_i^a$,$c_j^b$)\\
    $dHopCounter$[$c_j^b$] += dKGhops($c_i^a$,$c_j^b$)
  }
  }
}
\ForEach{$(c_i,score) \in dConnectCounter$}{
    $\mathcal{C} (c_i)$  = $dConnectCounter(c_i)/n$	  \tcp{Normalisation with respect to number of keywords spotted}
}
\ForEach{$(c_i,score) \in dHopCounter$}{
	$\mathcal{H} (c_i)$  = $dHopCounter(c_i)/n$
}
  \Return (Hop-Count $\mathcal{H}$, Connection-Count $\mathcal{C}$)

\caption{Connection Density}

\label{alg:conn}

\end{algorithm}

\vspace{-15pt}

\subsection{Adaptive E/R Learning} \label{ad-er}
\begin{figure}[t]
	\centering
	\includegraphics[width=.85\linewidth]{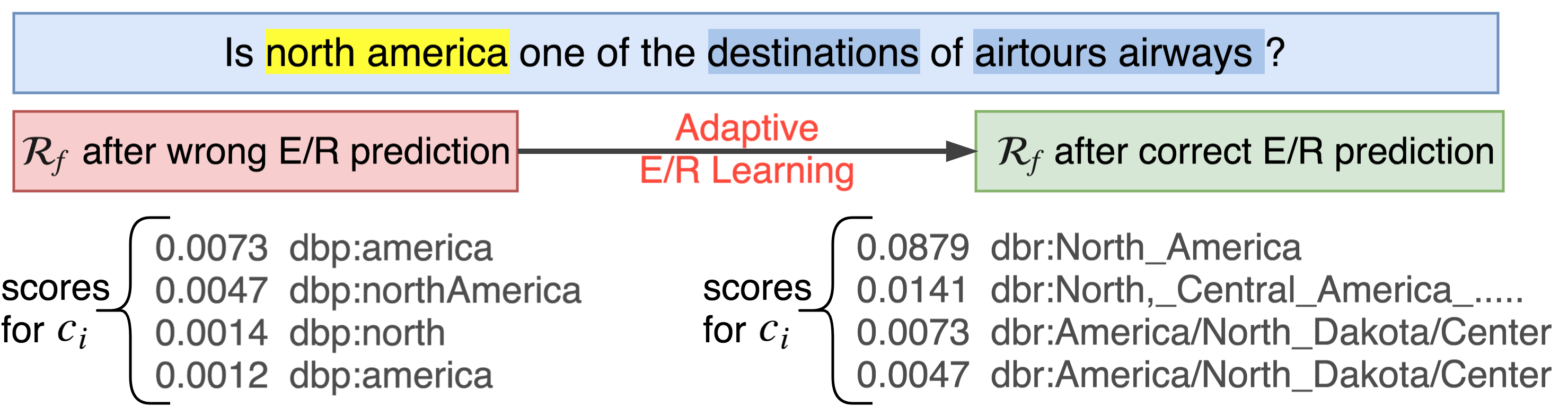}
	\vspace{-12pt}
	\caption{Adaptive E/R learning }
	\label{fig:adapER} 
	\vspace{-15pt}
\end{figure}

EARL uses a series of sequential modules with little to no feedback across them. Hence, the errors in one module propagate down the line. 
To trammel this, we implement an adaptive approach especially for curbing the errors made in the pre-processing modules. 
While conducting experiments, it was observed that most of the errors are in the shallow parsing phase, mainly because of grammatical errors in LC-QuAD which directly affects the consecutive E/R prediction and candidate selection steps.
If the E/R prediction is erroneous, it will search in a wrong Elasticsearch index for probable candidate list generation. 
In such a case none of the candidates $\in$ $c^i$ for a keyword would contain $\Bar{\textbf{c}}^i$ as is reflected by the probabilities assigned to $c^i$ by the re-ranker module. If the maximum probability assigned to $c^i$ is less than a very small threshold value, empirically chosen as 0.01, we re-do the steps from ER prediction after altering the original prediction. If the initial assigned probability is \textit{entity}, we change it to \textit{relation} and vice-versa, example figure \ref{fig:adapER}. 
This module is empirically evaluated in table \ref{EARLentityAcc}.%

\section{Evaluation}

\label{eval}

\textbf{Data Set}:
LC-QuAD~\cite{trivedi2017lc} is the largest complex questions data set available for QA over KGs. 
We have annotated this data set to create a gold label data set for entity and relation linking, i.e.~each question now contains the correct KG entity and relation URIs with their respective text spans in the question. 
This annotation was done in a semi-automated process and subsequently manually verified. 
The annotated dataset of 5000 questions is publicly available at \url{https://figshare.com/projects/EARL/28218}

\subsection{Experiment 1: Comparison of GTSP, LKH and Connection Density}
\textbf{Aim:} We evaluate hypotheses (\textbf{H1} and \textbf{H2}) that the connection density and GTSP can be used for joint linking task.
We also evaluate the LKH approximation solution of GTSP for doing this task.
We compare the time complexity of the three different approaches.
\textbf{Results:}  Connection density results in a similar accuracy as that of an exact GTSP solution with a better time complexity (see Table~\ref{exp0}). 
Connection density has worse time complexity than approximate GTSP solver LKH if we assume the best case of equal cluster sizes for LKH. 
However, it provides a better accuracy. 
Moreover, the average time taken in EARL using connection density (including the candidate generation step) is 0.42 seconds per question. 
\setlength{\tabcolsep}{3pt}
\begin{table}[b]
    \begin{center}
    \begin{tabular}{{l}{l}{l}{l}} \toprule 
\textbf{Approach} & \textbf{Accuracy (K=30)} & \textbf{Accuracy (K=10)}  & \textbf{Time Complexity} \\ 
\midrule 
Brute Force GTSP     & 0.61        &  0.62    & $\mathcal{O}(n^2 2^n)$ \\ 
LKH - GTSP           & 0.59        &  0.58    & $\mathcal{O}(nL^2)$ \\ 
Connection Density   & 0.61        &  0.62    & $\mathcal{O}(N^2L^2)$ \\ 
\bottomrule
    \end{tabular}
    \caption{Empirical comparison of Connection Density and GTSP: n = number of nodes in graph; L = number of clusters in graph; N = number of nodes per cluster; top K results retrieved from  ElasticSearch.}
    \label{exp0}
    \end{center}
   
    \vspace{-1cm}
\end{table}
Further observing Table~\ref{exp0}, we can see that the brute force GTSP solution and Connection Density have similar accuracy, but the brute force GTSP solution has exponential time complexity.
The approximate solution LKH has polynomial run time, but its accuracy drops compared to the brute force GTSP solution. 
Moreover, from a question answering perspective the ranked list offered by the Connection Density approach is useful since it can be presented to the user as a list of possible correct solutions or used by subsequent processing steps of a QA system. Hence, for further experiments in this section we used the connection density approach. 
\vspace{-0.5cm}
\subsection{Experiment 2: Evaluating Joint Connectivity and Re-ranker} 
\label{experi1}
\textbf{Aim}: Evaluating the performance of Connection Density for predicting the correct entity and relation candidates from a set of possible E-R candidates. 
Here we evaluate hypothesis \textbf{H2}, the correct candidates exhibit relatively dense and short-hop connections.

\vspace{-0.5cm}
\begin{table}
  \begin{center}   
\begin{tabular}{p{1.8cm}p{2.6cm}p{2.8cm}p{3.0cm}} \toprule 
\textbf{Value of k} & \textbf{$\mathcal{R}_f$ based on $\mathcal{R}_i$}  & \textbf{$\mathcal{R}_f$ based on $\mathcal{C},\mathcal{H}  $}  & \textbf{$\mathcal{R}_f$ based on $ \mathcal{R}_i, \mathcal{C},\mathcal{H}  $} \\ \midrule 

$k$ = 10      & 0.543     & 0.689         &  0.708    \\ 
$k$ = 30      & 0.544     & 0.666         &  0.735    \\ 
$k$ = 50     & 0.543     & 0.617         &  \textbf{0.737}    \\ 
$k$ = 100     & 0.540     & 0.534         &  0.733   \\ 
\midrule 
$k^*$ = 10      & 0.568     & 0.864        &  \textbf{0.905}     \\ 
$k^*$ = 30       & 0.554     & 0.779        &  0.864     \\ 
$k^*$ = 50      & 0.549     & 0.708        &  0.852     \\ 
$k^*$ = 50     & 0.545     & 0.603        &  0.817    \\ \bottomrule
    \end{tabular}
    \caption { Evaluation of joint linking performance}
    \label{exp1}
        \end{center}
    
    	\vspace{-1cm}
\end{table}
\vspace{-10pt}
\noindent 
\textbf{Metrics}: We use the mean reciprocal rank of the correct candidate $\Bar{\textbf{c}}^i$ for each entity/relation in the query. From the probable candidate list generation step, we fetch a list of top candidates for each identified phrase in a query with a $k$ value of  10, 30, 50 and 100, where k is the number of results from text search for each keyword spotted.
To evaluate the robustness of our classifier and features we perform two tests. i) On the top half of Table~\ref{exp1} we re-rank the top k candidates returned from the previous step. ii) On the bottom half of Table 4 we artificially insert the correct candidate into each list to purely test re-ranking abilities of our system (this portion of the table contains $k^*$ as the number of items in each candidate list). We inject the correct uris at the lowest rank (see $k^*$), if it was not retrieved in the top $k$ results from previous step. \\
\textbf{Results}: The results in Table~\ref{exp1} depict that our algorithm is able to successfully re-rank the correct URIs if the correct ones are already present. 
In case correct URIs were missing in the candidate list, we inserted URIs artificially as the last candidate . 
The MRR then increased from 0.568 to 0.905.

\subsection{Experiment 3: Evaluating Entity Linking}
\textbf{Aim}: To evaluate the performance of EARL with other state-of-the-art systems on the entity linking task. 
This also evaluates our hypothesis \textbf{H3}.\\ 
\textbf{Metrics}: We are reporting the performance on accuracy.
Accuracy is defined by the ratio of the correctly identified entities over the total number of entities present.\\ 
\textbf{Result}: EARL performs better entity linking than the other systems (Table~\ref{exp3}), namely Babelfy, DBpediaSpotlight, TextRazor and  AGDISTIS + FOX (limited to entity types - LOC, PER, ORG).
We conducted this test on the LC-QuAD and QALD-7 dataset\footnote{https://project-hobbit.eu/challenges/qald2017/}.
The value of \textbf{k} is set to 30 while re-ranking and fetching the most probable entity.
\begin{table}[t]
\centering 
    \begin{tabular}{p{5cm}p{3.5cm}p{3.5cm}} \toprule 
\textbf{System}          &\textbf{Accuracy LC-QuAD}     &\textbf{Accuracy - QALD} 
\\ \midrule
FOX~\cite{SPNG14FOX} + AGDISTIS~\cite{agdistis2014usbeck}                     & 0.36              & 0.30  \\
DBpediaSpotlight~\cite{dbpedia2011mendes}                    & 0.40              & 0.42  \\ 
TextRazor\footnote{~\url{https://www.textrazor.com/}}                   & 0.52             & 0.53  \\
Babelfy~\cite{babelfy}                    & 0.56             & 0.56  \\

 \midrule
EARL without adaptive learning     & 0.61               & 0.55  \\ 
EARL with adaptive learning         & \textbf{0.65}     &\textbf{0.57}
\\\bottomrule
    \end{tabular}
    \caption {Evaluating EARL's Entity Linking performance}
    \label{exp3}
    \vspace{-1cm}
\label{EARLentityAcc}
\end{table}

\subsection{Experiment 4: Evaluating Relation Linking}
\textbf{Aim}:
Given a question, the task is to the perform relation linking in the question. This also evaluates our hypothesis \textbf{H3}.\\
\textbf{Metrics}: We use the same accuracy metric as in the Experiment 3\\
\textbf{Results}: As reported in Table~\ref{exp4}, EARL outperforms other approaches we could run on LC-QuAD and QALD. 
The large difference in accuracy of relation-linking over LC-QuAD over QALD, is due to the face that LC-QuAD has 82\% questions with more than one relation, thus detecting relation phrases in the question was difficult. 
\vspace{-10pt}
\begin{table}
\centering 
     \begin{tabular}{p{4.8cm}p{3.2cm}p{3.2cm}} 
    \toprule 
    \textbf{System}          &\textbf{Accuracy LC-QuAD}     &\textbf{Accuracy - QALD} \\ \midrule
    ReMatch~\cite{RelMatch}                     & 0.12              & 0.31 \\
    RelMatch~\cite{ReMatchKuldeep}                    & 0.15              & 0.29  \\ 
    \midrule
    EARL without adaptive learning     & 0.32               & 0.45  \\ 
    EARL with adaptive learning         & \textbf{0.36}     &\textbf{0.47} 
    \\\bottomrule
    \end{tabular}
    \caption {Evaluating EARL's Relation Linking performance}
    \label{exp4}
   \vspace{-1cm}
\label{EARLRelAcc}
\end{table}
\vspace{-15pt}

\section{Discussion}
%

Our analysis shows that we have provided two tractable (polynomial with respect to the number of clusters and the elements per cluster) approaches of solving the joint entity and relation linking problem. 
We experimently achieve similar accuracy as the exact GTSP solution with both LKH-GTSP and Connection Density with better time complexity, which allows us to use the system in QA engines in practice. It must be noted that one of the salient features of LKH-GTSP is that it requires no training data for the disambiguation module while on the other hand Connection Density performs better given training data for its XGBoost classifier. 
 While the system was tested on DBpedia, it is not restricted to a particular knowledge graph. 

There are some limitations: 
The current approach does not tackle questions with hidden relations, such as "How many shows does HBO have?". Here the semantic understanding of the corresponding SPARQL query is to count all TV shows (\textit{dbo:TelevisionShow}) which are owned by (\textit{dbo:company}) the HBO (\textit{dbr:HBO}). Here \textit{dbo:company} is the hidden relation which we do not attempt to link. However, it could be argued that this problem goes beyond the scope of relation linking and could be better handled by the query generation phase of a semantic QA system. 

Another limitation is that EARL cannot be used as inference tool for entities as required by some questions. For example Taikonaut is an astronaut with Chinese nationality. The system can only link taikonaut to \textit{dbr:Astronaut}, but additional information can not be captured. It should be noted, however, that EARL can tackle the problem of the "lexical gap" to a great extent as it uses synonyms via the grammar inflection forms.

Our approaches of LKH-GTSP and Connection Density both have polynomial and approximately similar time complexities. 
EARL with either Connection Density or LKH-GTSP can process a question in a few hundred milliseconds on a standard desktop computer on average. 
The result logs, experimental setup and source code of our system are publicly available at: \href{https://github.com/AskNowQA/EARL}{https://github.com/AskNowQA/EARL}.
\vspace{-10pt}
\section{Conclusions and Future Work}

Here we propose EARL, a framework for joint entity and relation linking. 
We provided two strategies for joint linking - one based on reducing the problem to an instance of the Generalised Travelling Salesman problem and the other based on a connection density based machine learning approach. 
Our experiments on QA benchmarks resulted in accuracies which are significantly above the results of current state-of-the-art approaches for entity and relation linking. In future, we will improve the candidate generation phase to ensure that a higher proportion of correct candidates are retrieved. 

\textbf{Acknowledgement}
This work is supported by the funding received from the EU H2020 projects WDAqua (ITN, GA. 642795) and HOBBIT (GA. 688227). 

\bibliographystyle{abbrv}
\bibliography{bibliography}

\end{document}